%% file: acl2016.tex
\documentclass[11pt]{article}
\usepackage{acl2016}
\usepackage{times}
\usepackage{url}
\usepackage{latexsym}
\usepackage[draft]{hyperref}
\usepackage{wrapfig}
\usepackage{lipsum}
\usepackage{bbm}
\usepackage{array,multirow,graphicx,xcolor}
\usepackage{algorithm}
\usepackage[noend]{algpseudocode}
\usepackage{amsmath}
\usepackage{lipsum}
\usepackage{tabularx}
\usepackage{booktabs}
\usepackage{ragged2e}
\usepackage{float}
\usepackage[font=small]{caption}
\usepackage{array}
\newcolumntype{L}{>{\centering\arraybackslash}m{3cm}}

\newcommand{\h}{{\bf h}}
\newcommand{\w}{{\bf w}}

\newcommand{\z}{{\bf z}}

\newcommand{\LSSVM}{{ LSSVM}}

\newcolumntype{Y}{>{\RaggedRight\arraybackslash}X}

\aclfinalcopy % Uncomment this line for the final submission
\title{Science Question Answering using Instructional Materials}

\author{{\bf Mrinmaya Sachan} \hspace{1.5cm} {\bf Avinava Dubey} \hspace{1.5cm} {\bf Eric P. Xing}\\
School of Computer Science\\  
Carnegie Mellon University\\
  {\tt \{mrinmays, akdubey, epxing\}@cs.cmu.edu}\\
}

\date{}

\begin{document}
\maketitle

\input{abstract}

\input{introduction}

\input{method}

\input{experiments}

\input{conclusions}

\bibliography{acl2016}
\bibliographystyle{acl2016}

\end{document}

%% file: abstract.tex
\begin{abstract}
We provide a solution for elementary science tests using instructional materials. We posit that there is a hidden structure that explains the correctness of an answer given the question and instructional materials and present a unified max-margin framework that learns to find these hidden structures (given a corpus of question-answer pairs and instructional materials), and uses what it learns to answer novel elementary science questions. 
Our evaluation shows that our framework outperforms several strong baselines.
\end{abstract}

%% file: introduction.tex
\section{Introduction}
%Standardized tests have often been proposed as ``drivers for progress in AI'' \cite{Clark:2016mag}. These include tests on understanding passages and answering questions about them \cite{richardson:2013}, algebra word problems \cite{kushman:2014}, geometry problems \cite{seo:2014}, etc. Many of these tests require sophisticated understanding and aim to push the boundaries of AI. 
%Besides they are accessible, easily comprehensible, clearly measurable, and offer a graduated progression from simple tasks to those requiring deep understanding of the world.
%{\bf Elementary science tests:}
%Elementary Science tests \cite{Clark:2015} offer another interesting and challenging opportunity.
%As discussed in \newcite{Clark:2015}, \newcite{Clark:2016mag}, t
%These tests are quite challenging because of a wide variety of knowledge and reasoning required to answer these questions.
% Despite significant interest in the recent years \cite{khot:2015,li:2015,Clark:2016}, the problem remains unsolved.
%{\bf Answer-entailing structures:}

\begin{figure*}
\begin{minipage}[c]{0.6\textwidth}
\includegraphics[width=\textwidth]{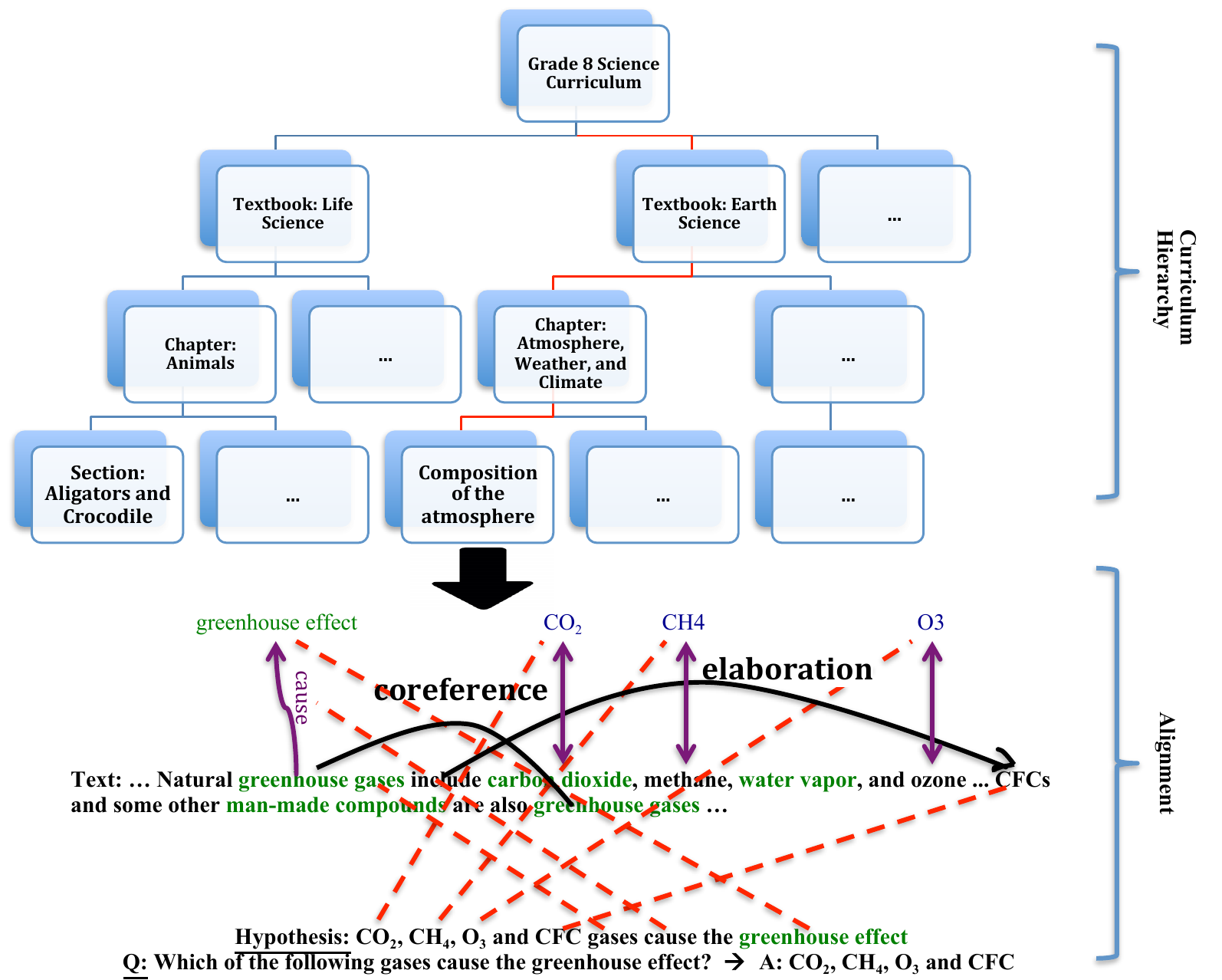}
\end{minipage}\hfill
\begin{minipage}[c]{0.395\textwidth}
\caption{\footnotesize{An example \textit{answer-entailing structure}. The answer-entailing structure consists of selecting a particular textbook from the curriculum, picking a chapter in the textbook, picking a section in the chapter, picking sentences in the section and then aligning words/mwe's in the hypothesis (formed by combining the question and an answer candidate) to words/mwe's in the picked sentences or some related ``knowledge'' appropriately chosen from additional knowledge stores. In this case, the relation (greenhouse gases, cause, greenhouse effect) and the equivalences (e.g. carbon dioxide = $CO_2$) -- shown in violet -- are hypothesized using external knowledge resources. The dashed red lines show the word/mwe alignments from the hypothesis to the sentences (some word/mwe are not aligned, in which case the alignments are not shown), the solid black lines show coreference links in the text and the RST relation (elaboration) between the two sentences. The picked sentences do not have to be contiguous sentences in the text. All mwe's are shown in green.}} \label{structure-fig}
\end{minipage}
\end{figure*}

%In this paper, w
We propose an approach for answering multiple-choice elementary science tests \cite{Clark:2015}
using the science curriculum of the student and other domain specific knowledge resources. Our approach learns latent \textit{answer-entailing structures} that align question-answers with appropriate snippets in the curriculum.
The student curriculum usually comprises of a set of textbooks. Each textbook, in-turn comprises of a set of chapters, each chapter is further divided into sections -- each discussing a particular science concept. Hence, the answer-entailing structure consists of selecting a particular textbook from the curriculum, picking a chapter in the textbook, picking a section in the chapter, picking a few sentences in the section and then aligning words/multi-word expressions (mwe's) in the hypothesis (formed by combining the question and an answer candidate) to a words/mwe's in the picked sentences. The answer-entailing structures are further refined using external domain-specific knowledge resources such as science dictionaries, study guides and semi-structured tables (see Figure \ref{structure-fig}). These domain-specific knowledge resources can be very useful forms of knowledge representation as shown in previous works \cite{Clark:2016}.
% and our own experiments.

Alignment is a common technique in many NLP applications such as MT \cite{blunsom:2006}, RTE \cite{Sammons:09,maccartney:2008,Yaox:2013,Sultan:14}, QA \cite{Berant:13,Yih:2013,Yao:2014,sachan:2015}, etc.
Yet, there are three key differences between our approach and alignment based approaches for QA in the literature: (i) We incorporate the curriculum hierarchy (i.e. the book, chapter, section bifurcation) into the latent structure. This helps us jointly learn the retrieval and answer selection modules of a QA system. Retrieval and answer selection are usually designed as isolated or loosely connected components in QA systems \cite{ferrucci:2012} leading to loss in performance -- our approach mitigates this shortcoming. 
(ii) Modern textbooks typically provide a set of review questions after each section to help students understand the material better. We make use of these review problems to further improve our model. These review problems have additional value as part of the latent structure is known for these questions.
(ii) We utilize domain-specific knowledge sources such as study guides, science dictionaries or semi-structured knowledge tables within our model.

The joint model is trained in max-margin fashion using a latent structural SVM (LSSVM) where the answer-entailing structures are latent.
We train and evaluate our models on a set of $8^{th}$ grade science problems, science textbooks and multiple domain-specific knowledge resources. We achieve superior performance vs.\ a number of baselines.
%We show value in using review questions.
% and release an easy-to-use form of these questions for our dataset.

%% file: method.tex
\section{Method}
{\bf Science QA as Textual Entailment:} 
%Let us formalize the science QA setting. 
First, we consider the case when review questions are not used. For each question $q_i \in Q$, let $A_i = \{ a_{i1}, \ldots, a_{im}\}$ be the set of candidate answers to the question
%We treat the problem of answering multiple-choice science tests using academic curriculum as a problem of selecting the best answer from a set of candidate answers. 
\footnote{Candidate answers may be pre-defined, as in multiple-choice QA, or may be undefined but easy to extract with a degree of confidence (e.g., by using a pre-existing system)}.
We cast the science QA problem as a textual entailment problem by converting each question-answer candidate pair $(q_i, a_{i,j})$ into a hypothesis statement $h_{ij}$ (see Figure \ref{structure-fig})% For example, the question ``Which of the following gases cause the greenhouse effect?'' and answer candidate ``$CO_2$, $CH_4$, $O_3$ and CFC'' in Figure \ref{structure-fig} are combined to achieve a hypothesis ``$CO_2$, $CH_4$, $O_3$ and CFC gases cause the greenhouse effect''
\footnote{We use a set of question matching/rewriting rules to achieve this transformation. The rules match each question into one of a large set of pre-defined templates and applies a unique transformation to the question \& answer candidate to achieve the hypothesis. Code provided in the supplementary.}. For each question $q_i$, the science QA task thereby reduces to picking the hypothesis $\hat h_i$ that has the highest likelihood of being entailed by the curriculum among the set of hypotheses $\h_i = \{h_{i1},\ldots,h_{im}\}$ generated for that question. Let $h^{*}_{i} \in \h_i$ be the correct hypothesis corresponding to the correct answer.
% Now let us define latent answer-entailing structures.
\\
{\bf Latent Answer-Entailing Structures}
help the model in providing evidence for the correct hypothesis.
% We consider the quality of a one-to-one word alignment from a hypothesis to snippets in the textbooks as a proxy for the evidence.
As described before, the structure depends on: (a) snippet from the curriculum hierarchy chosen to be aligned to the hypothesis, (b) external knowledge relevant for this entailment, and (c) the word/mwe alignment. The snippet from the curriculum to be aligned to the hypothesis is determined by walking down the curriculum hierarchy and then picking a set of sentences from the section chosen. Then, a subset of relevant external knowledge in the form of triples and equivalences (called knowledge bits) is selected from our reservoir of external knowledge (science dictionaries, cheat sheets, semi-structured tables, etc). Finally, words/mwe's in the hypothesis are aligned to words/mwe's in the snippet or knowledge bits. Learning these alignment edges helps the model determine which semantic constituents should be compared to each other. These alignments are also used to generate more effective features.
The choice of snippets, choice of the relevant external knowledge and the alignments in conjunction form the latent answer-entailing structure. Let $\z_{ij}$ represent the latent structure for the question-answer candidate pair $(q_i, a_{i,j})$.
%Furthermore, let $z_{ij}^1$, $z_{ij}^2$, $z_{ij}^3$, $z_{ij}^4$, $z_{ij}^5$ represent the choice of the text book in the curriculum, choice of a chapter from the textbook ($z_{ij}^1$), a section from the chapter ($z_{ij}^2$), a set of sentences in section ($z_{ij}^3$) and an alignment of hypothesis $h_{ij}$ on the set of sentences picked ($z_{ij}^4$), respectively.
\\
{\bf Max-Margin Approach:}
We treat science QA as a structured prediction problem of ranking the hypothesis set $\h_i$ such that the correct hypothesis is at the top of this ranking. 
%Let us assume that the set of hypothesis set is $H = \{ \h_1,\h_2,...\h_N\}$ where $\h_i \in H$ is the set of hypothesis corresponding to query $q_i$. Dropping the subscript $i$ for clarity we will use the hypothesis set $\h \in H$ for our formulation and the corresponding latent structure using $\z$. 
We learn a scoring function $S_{\w}(h,\z)$ with parameter $\w$ such that the score of the correct hypothesis $h_i^*$ and the corresponding best latent structure $\z_i^*$ is higher than the score of the other hypotheses and their corresponding best latent structures. In fact, in a max-margin fashion, we want that $ S_{\w}(h_i^*,\z_i^*) > S(h_{ij},\z_{ij}) + 1 - \xi_i$ for all $h_j \in \h\setminus h^*$ for some slack $\xi_i$. Writing the relaxed max margin formulation:

{\small
\begin{eqnarray*}
\min_{||\w||} \frac{1}{2}||\w||^2_2 +
C \sum_i \max\limits_{\z_{ij}, h_{ij} \in \h_i \setminus h_i^*} S_{\w}(h_{ij},\z_{ij}) + \Delta(h_i^*, h_{ij}) \\
- C \sum_i S_{\w}(h_i^*,\z_i^*)\quad\quad\quad (1)\hspace{2.36cm}
\end{eqnarray*}
}
We use 0-1 cost, i.e. $\Delta(h_i^*, h_{ij}) = \mathbbm{1}(h_i^* \neq h_{ij})$
If the scoring function is convex then this objective is in concave-convex form and hence can be solved by the concave-convex programming procedure (CCCP) \cite{Yuille:2003}. We assume the scoring function to be linear:$
S_{\w}(h,\z) = {\w}^T\psi(h,\z)$.
Here, $\psi(h,\z)$ is a feature map discussed later.
% Efficient but involved solutions using the cutting-plane algorithm \cite{Yu/Joachims/09a} is known for this problem. %However, this relies on solving the inference problem exactly. This is not possible in our formulation as the space of latent structures is very large. Hence, we resort to an alternate minimization procedure described below. To solve equation 1 we can 
The CCCP algorithm essentially alternates between solving for $\z_i^*$, $\z_{ij}\ \forall j$ s.t. $ h_{ij} \in \h_i \setminus h_i^*$ and $\w$ to achieve a local minima. In the absence of information regarding the latent structure $\z$ we pick the structure that gives the best score for a given hypothesis i.e. $\arg \max_{z} S_{\w}(h,z)$.
% Let for hypothesis set $\h$ with $h^*$ being the correct hypothesis have $\z^* = \arg \max_{\z} S_w(h^*,\z)$. 
The complete procedure is given in the supplementary.\\
{\bf Inference and knowledge selection:} We use beam search with a fixed beam size (5) for inference. We infer the textbook, chapter, section, snippet and alignments one by one in this order. In each step, we only expand the five most promising (given by the current score) substructure candidates so far. During inference, we select top 5 knowledge bits (triples, equivalences, etc.) from the knowledge resources that could be relevant for this question-answer. This is done heuristically by picking knowledge bits that explain parts of the hypothesis not explained by the chosen snippets.
%\begin{algorithm}
%\caption{Alternate Minimization for LSSVM}\label{algo_SSVM1}
%\begin{algorithmic}[1]
%\State Initialise $\w$
%\State $C_i = \emptyset$ $\forall i = 1\ldots n$
%\Repeat
%\For{ $i =1,\ldots ,n$}
%\State $\z_i^* =\argmax\limits_{\z} S_{\w} (h_i^*,\z)$
%\For{$h_{ij} \in \h_i \setminus h_i^* $}
%\State $\z_{ij} = \argmax\limits_{\z} S_{\w} (h_{ij},\z)$
%\EndFor
%\State $h_{i}^t,\z_{i}^t = \argmax\limits_{h_{ij}\neq h_i^*,\z_{ij}}S_{\w}(h_{ij},\z_{ij})$
%\State $C_i = C_i \cup ( \{h_{i}^t,z_i^t\} \cap (S_{\w}(h_{ij},\z_{ij}) > S_{\w}(h_i^*,\z_i^*)-1) )$
%\EndFor
%\State {\bf Solve QP:} {\small
%\begin{eqnarray*}
%&&\min_{\w} \frac{1}{2}||\w||^2_2 + \sum_i \xi_i \\
%&&\mbox{s.t.} S_{\w}(h_i^*,\z_i^*) > S_{\w}(h,\z) + 1 - \xi_i \ \ \forall \{h,z\} \in C_i\\
%&&\hspace{5.32cm} \forall i = 1\ldots n
%\end{eqnarray*} }
%\Until{ Convergence}
%\end{algorithmic}
%\end{algorithm}
\\
{\bf Incorporating partially known structures:} Now, we describe how review questions can be incorporated.
As described earlier, modern textbooks often provide review problems at the end of each section.
% These review problems help students review and better understand the material. Inspired by this, we make use of these review problems along with the standardized test problems to improve our model. 
These review problems have value as part of the answer-entailing structure (textbook, chapter and section) is known for these problems.
%Hence, we now have two sets of questions $\mathcal{Q}_{kaggle}$ and $\mathcal{Q}_{review}$. For the hypotheses derived from the review questions, the answer entailing structure is partially known. Let the set of hypothesis corresponding to $\mathcal{Q}_{kaggle}$ be $\mathcal{H}_{kaggle}$ and the set of hypotheses corresponding to $\mathcal{Q}_{review}$ be $\mathcal{H}_{review}$. 
In this case, we use the formulation (equation 1) except that the max over $\z$ for the review questions is only taken over the unknown part of the latent structure.
\\
{\bf Multi-task Learning:} Question analysis is a key component of QA systems. Incoming questions are often of different types (counting, negation, entity queries, descriptive questions, etc.). Different types of questions usually require different processing strategies.
% Yet, they may share many properties.
%\avicomment{write a bit about how it is useful}
Hence, we also extend of our LSSVM model to a multi-task setting where each question $q_i$ now also has a pre-defined associated type $t_i$ and each question-type is treated as a separate task. Yet, parameters are shared across tasks,which allows the model to exploit the commonality among tasks when required. %This calls for a multi-task LSSVM (MTLSSVM). 
We use the MTLSSVM formulation from
% We assume that each hypothesis set $\h_i$ now also has an associated type $t_i$. 
%We assume that the task specific weight vector $\w_t$ factorizes as a globally shared weight vector $\w$ and a task specific weight vector $\bv_t$.
%We again assume the scoring function to be a linear of the form $S_{\w,\bv}(h_i,\z_i,t_i) = (\w + \bv_{t_i})^T \psi(h_i,\z_i)$. 
%With this assumption, we can use the trick described in \newcite{Evgeniou:2004} to solve the MTLSSVM where we redefine the feature map as:
%\begin{align}\nonumber
%\Phi(h_{ij,}\z_{ij},t_i=s) &= (\frac{\psi(h_i,\z_i)}{\mu}, \underbrace{{\bf 0,\ldots,0}}_{s-1},\\
%\nonumber &\quad \quad \psi(h_i,\z_i),\underbrace{ {\bf 0,\ldots,0}}_{S-s})
%\end{align}
%(for $\mu$ tuned on the dev split) and redefine the scoring function as $ S_{W}(h_i,\z_i,t_i) = W^T\Phi(h_{ij,}\z_{ij},t_i)$. 
\newcite{Evgeniou:2004} which was also used in a reading comprehension setting by \newcite{sachan:2015}.
In a nutshell, the approach redefines the LSSVM feature map and shows that the MTLSSVM objective takes the same form as equation 1 with a kernel corresponding to the feature map. Hence, one can simply redefine the feature map and reuse LSSVM algorithm to solve the \textit{MTLSSVM}.
\\
{\bf Features:}
%Now we define the feature function $\psi (h ,\z)$. Here the hypothesis $h$ was formed by combining the question $q$ and answer candidate $a$ as described earlier, $\z$ being the answer-entailing structure also defined earlier. 
Our feature vector $\psi (h ,\z)$ decomposes into five parts, where each part corresponds to a part of the answer-entailing structure.
% The first three parts incorporate IR style features. 
For the first part, we index all the textbooks and score the top retrieved textbook by querying the hypothesis statement. We use tf-idf and BM25 scorers resulting in two features. Then, we find the jaccard similarity of bigrams and trigrams in the hypothesis and the textbook to get two more features for the first part. Similarly, for the second part we index all the textbook chapters and compute the tf-idf, BM25 and bigram, trigram features. For the third part we index all the sections instead.
%The features for the fourth and fifth parts are inspired from \newcite{sachan:2015}.
The fourth part has features based on the text snippet part of the answer-entailing structure. Here we do a deeper linguistic analysis and include features for matching local neighborhoods in the snippet and the hypothesis: features for matching bigrams, trigrams, dependencies, semantic roles, predicate-argument structure as well as the global syntactic structure: a tree kernel for matching dependency parse trees of entire sentences \cite{srivastava:13}. If a text snippet contains the answer to the question, it should intuitively be similar to the question as well as to the answer. Hence, we add features that are the element-wise product of features for the text-question match and text-answer match.
% In addition to features for the exact word/phrase match of the snippet and the hypothesis, we also add features using two paraphrase databases: ParaPara \cite{chan:2011} and DIRT \cite{lin:2001}. These databases contain paraphrase rules of the form string$_1 \rightarrow$ string$_2$. ParaPara has rules like ``imprisoned'' $\rightarrow$ ``sent to jail'', ``huge amount of'' $\rightarrow$ ``large quantity of'', etc. extracted through bilingual pivoting and DIRT database contains rules like ``A is the author of B'' $\rightarrow$ ``A wrote B'', ``A caused B'' $\rightarrow$ ``B is triggered by A'', etc. extracted using the distributional hypothesis \cite{harris:54}. Whenever we have a substring in the text snippet that can be transformed into another using any of these two databases, we keep match features for the substring with a higher score (according to the current $\w$) and ignore the other substring.
Finally, we also have features corresponding to the RST \cite{mann:1988} and coreference links to enable inference across sentences. RST tells us that sentences with discourse relations are related to each other and can help us answer certain kinds of questions \cite{Jansen:2014}.
For example, the ``cause'' relation between sentences in the text can often give cues that can help us answer ``why'' or ``how'' questions. 
Hence, we add additional features - conjunction of the rhetorical structure label from a RST parser and the question word - to our feature vector. Similarly, the entity and event co-reference relations allow us to reason about repeating entities or events. Hence, we replace an entity/event mention with their first mentions if that results into a greater score.
% into We add additional features of the aforementioned types by replacing entity/event mentions with their first mentions.
For the alignment part, we induce features based on word/mwe level similarity of aligned words: (a) Surface-form match (Edit-distance), and (b) Semantic word match (cosine similarity using SENNA word vectors \cite{collobert:2011} and ``Antonymy'' `Class-Inclusion' or `Is-A' relations using Wordnet). Distributional vectors for mwe's are obtained by adding the vector representations of comprising words \cite{Mitchell:2008}. To account for the hypothesized knowledge bits, whenever we have the case that a word/mwe in the hypothesis can be aligned to a word/mwe in a hypothesized knowledge bit to produce a greater score, then we keep the features for the alignment with the knowledge bit instead.
\\
{\bf Negation}
%We empirically found that a key limitation in our formulation is its inability to handle negation. Negation is especially hurtful to our model as it not only results in poor performance on questions that require us to reason with negated facts, it provides our model with a wrong signal (facts usually align well with their negated versions). We use a simple heuristic to overcome the negation problem. We detect negation (either in the hypothesis or a sentence in the text snippet aligned to it) using a small set of manually defined rules that test for presence of words such as ``not'', ``n't'', etc. Then, we flip the partial order - i.e. the correct hypothesis is now ranked below the other competing hypotheses. For inference at test time, we also invert the prediction rule i.e. we predict the hypothesis (answer) that has the least score under the model.
%We empirically found that n
Negation is a concern for our approach as facts usually align well with their negated versions.
% Negation questions not only reduces the accuracy of our model but these questions also provide incorrect signal to the learning algorithm as facts usually align well with their negated versions. 
To overcome this, we use a simple heuristic. During training, if we detect negation using a set of simple rules that test for the presence of negation words (``not'', ``n't'', etc.), we flip the partial order adding constraints that require that the correct hypothesis to be ranked below all the incorrect ones. During test phase if we detect negation, we predict the answer corresponding to the hypothesis with the lowest score.

%% file: experiments.tex
\section{Experiments}
\begin{figure}[t]
\centering
\includegraphics[scale = 0.27]{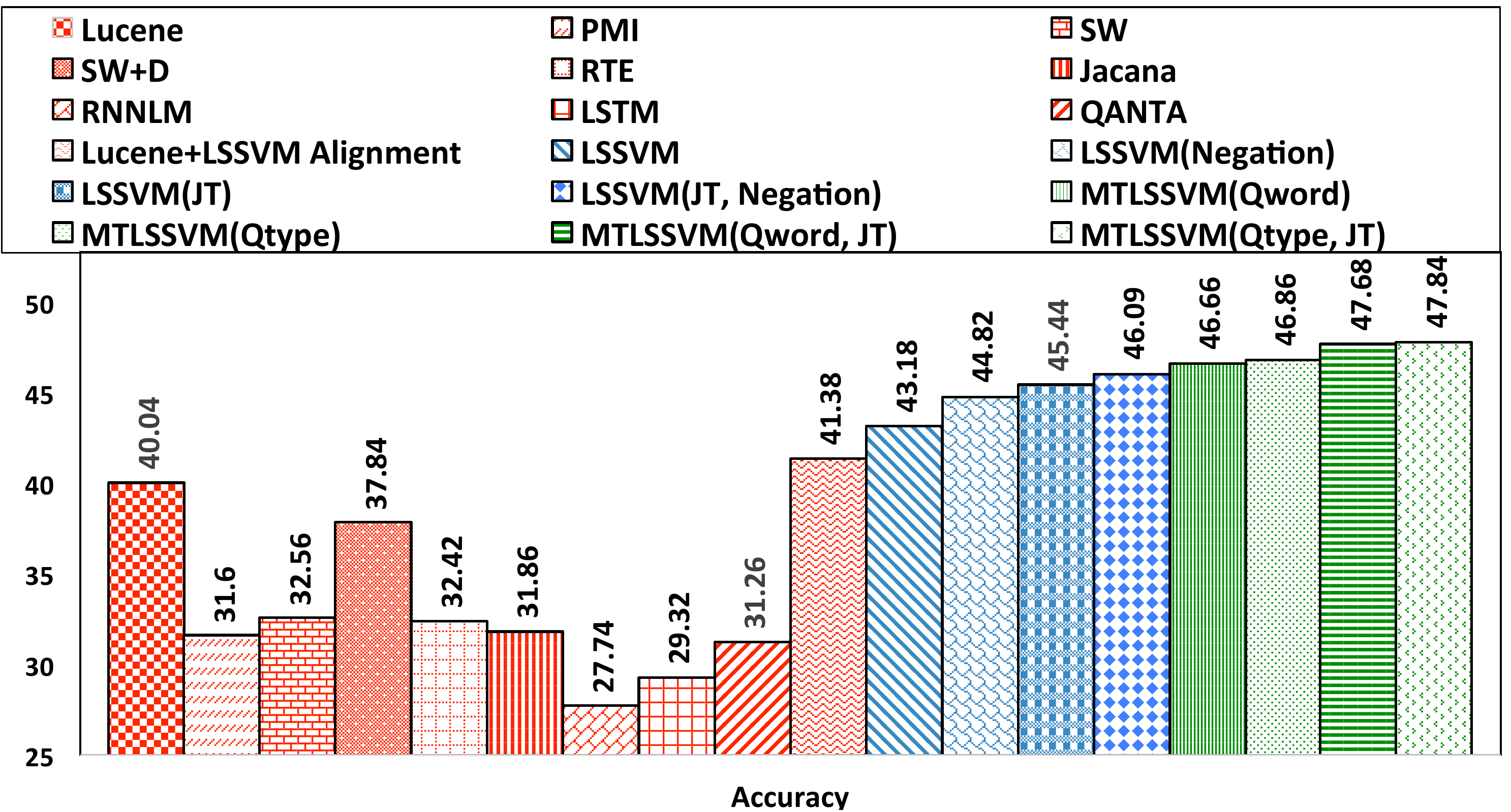}
\caption{\footnotesize{Variations of our method vs several baselines on the Science QA dataset. Differences between the baselines and \LSSVM s, the improvement due to negation, the improvements due to multi-task learning and joint-learning are significant ($p<0.05$) using the two-tailed paired T-test.}\label{result-fig}}\vspace{-0.2cm}
\end{figure}

\begin{figure}
\centering
\includegraphics[scale = 0.28]{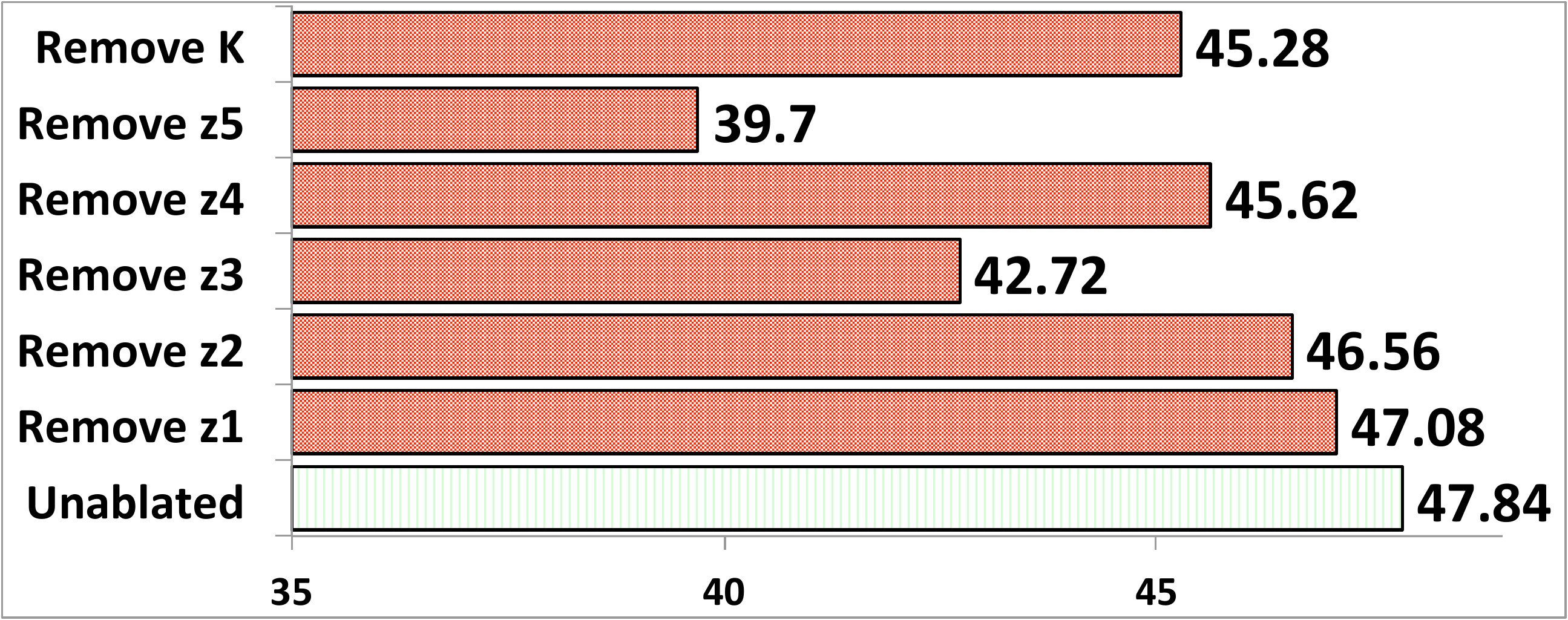}
\caption{\footnotesize{Ablation on \textit{MTLSSVM(Qword, JT) model}}\label{ablation-fig}}
\end{figure}

{\bf Dataset:}
We used a set of $8^{th}$ grade science questions released as the training set in the Allen AI Science Challenge\footnote{\href{https://www.kaggle.com/c/the-allen-ai-science-challenge/}{https://www.kaggle.com/c/the-allen-ai-science-challenge/}} for training and evaluating our model. The dataset comprises of 2500 questions. Each question has 4 answer candidates, of which exactly one is correct. We used questions 1-1500 for training, questions 1500-2000 for development and questions 2000-2500 for testing. We also used publicly available $8^{th}$ grade science textbooks available through $\href{http://www.ck12.org}{ck12.org}$. The science curriculum consists of seven textbooks on Physics, Chemistry, Biology, Earth Science and Life Science. Each textbook on an average has 18 chapters, and each chapter in turn is divided into 12 sections on an average.
Also, as described before, each section, on an average, is followed by 3-4 multiple choice review questions (total 1369 review questions).
% We collect and release them in an easy-to-use form.
%, meant to help students better understand the content properly. 
%These questions can benefit our model as the latent structure is partially known for these problems. To test this, we consider jointly training (JT) our model using review questions as well.
%We also provide results in both scenarios -- when kaggle/review questions are also used for training, or not. 
%We have a total of 1369 review questions.
%We split the review questions into training, dev and test splits in approximately the same proportion (60\%, 20\%, 20\%).
We collected a number of domain specific science dictionaries, study guides, flash cards and semi-structured tables (Simple English Wiktionary and Aristo Tablestore) available online and create triples and equivalences used as external knowledge.
\\
{\bf Baselines:}
\begin{table}[t!]
\scriptsize
\begin{tabular}{|l|l|}
%\begin{tabularx}{|@{} l | Y @{}|} % use 'Y' for first column
\hline
{\bf Question Category} & {\bf Example} \\
\hline\hline
\multicolumn{1}{|m{1.93cm}|}{Questions without context:} & \multicolumn{1}{m{5.65cm}|}{Which example describes a learned behavior in a dog?} \\ \hline
\multicolumn{1}{|m{1.93cm}|}{Questions with context:} & \multicolumn{1}{m{5.65cm}|}{When athletes begin to exercise, their heart rates and respiration rates increase. At what level of organization does the human body coordinate these functions?} \\ \hline
\multicolumn{1}{|m{1.93cm}|}{Negation Questions:}&\multicolumn{1}{m{5.65cm}|}{A teacher builds a model of a hydrogen atom. A red golf ball is used for a proton, and a green golf ball is used for an electron. Which is not accurate concerning the model?}\\\hline
\end{tabular}
\caption{\footnotesize{Example questions for \textit{Qtype} classification}} 
\label{table:qclass}
\end{table}
We compare our framework with ten baselines.
The first two baselines (\textit{Lucene} and \textit{PMI}) are taken from \newcite{Clark:2016}. The \textit{Lucene} baseline scores each answer candidate $a_i$ by searching for the combination of the question $q$ and answer candidate $a_i$ in a lucene-based search engine and returns the highest scoring answer candidate. The \textit{PMI} baseline similarly scores each answer candidate $a_i$ by computing the point-wise mutual information to measure the strength of the association between parts of the question-answer candidate combine and parts of the CK12 curriculum.
The next three baselines, inspired from \newcite{richardson:2013}, retrieve the top two CK12 sections querying $q+a_i$ in \textit{Lucene} and score the answer candidates using these documents. The \textit{SW} and \textit{SW+D} baselines match bag of words constructed from the question and the answer answer candidate to the retrieved document.
%Since this ignores long range dependencies, the \textit{SW+D} baseline accounts for intra-word distances as well. 
The \textit{RTE} baseline uses textual entailment \cite{stern:2012} to score answer candidates as the likelihood of being entailed by the retrieved document.
Then we also tried other approaches such as the \textit{RNN} approach described in \newcite{Clark:2016}, \textit{Jacana aligner} \cite{Yaox:2013} and two neural network approaches, \textit{LSTM} \cite{hochreiter:1997} and \textit{QANTA} \cite{Iyyer:2014} %described in \newcite{sachan:2015}. 
They form our next four baselines.
To test if our approach indeed benefits from jointly learning the retrieval and the answer selection modules, our final baseline \textit{Lucene+LSSVM Alignment}
%applies latent structural SVM in two parts. First, it 
retrieves the top section by querying $q+a_i$ in \textit{Lucene} and then learns the remaining answer-entailment structure (alignment part of the answer-entailing structure in Figure \ref{structure-fig}) using a LSSVM.
%\footnote{One can think of this baseline as stacking \newcite{sachan:2015} on top of \textit{Lucene}}
\\
{\bf Task Classification for Multitask Learning:}
We explore two simple question classification schemes. The first classification scheme classifies questions based on the question word (what, why, etc.). We call this \textit{Qword} classification.
The second scheme is based on the type of the question asked and classifies questions into three coarser categories: (a) questions without context, (b) questions with context and (c) negation questions. This classification is based on the observation that many questions lay down some context and then ask a science concept based on this context. However, other questions are framed without any context and directly ask for the science concept itself. Then there is a smaller, yet, important subset of questions that involve negation that also needs to be handled separately. Table \ref{table:qclass} gives examples of this classification. We call this classification \textit{Qtype} classification\footnote{We wrote a set of question matching rules (similar to the rules used to convert question answer pairs to hypotheses) to achieve this classification}.
\\
{\bf Results:}
We compare variants of our method\footnote{We tune the SVM regularization parameter $C$ on the development set. We use Stanford CoreNLP, the HILDA parser \cite{feng-hirst:2014}, and jMWE \cite{kulkarni:2011} for linguistic preprocessing} where we consider our modification for negation or not and multi-task LSSVMs. We consider both kinds of task classification strategies and joint training (JT). Finally, we compare our methods against the baselines described above. We report accuracy (proportion of questions correctly answered) in our results.
%and NDCG$_4$ \cite{jarvelin:2002}. Unlike classification accuracy which evaluates if the prediction is correct or not, NDCG$_4$, being a measure of ranking quality, evaluates the position of the correct answer in our predicted ranking.
Figure \ref{result-fig} shows the results.
% on the kaggle questions and the review questions. 
First, we can immediately observe that all the LSSVM models have a better performance than all the baselines. We also found an improvement when we handle negation using the heuristic described above\footnote{We found that the accuracy over test questions tagged by our heuristic as negation questions went up from 33.64 percent to 42.52 percent and the accuracy over test questions not tagged as negation did not decrease significantly}. MTLSSVMs showed a boost over single task LSSVM. \textit{Qtype} classification scheme was found to work better than \textit{Qword} classification which simply classifies questions based on the question word.
The multi-task learner could benefit even more if we can learn a better separation between the various strategies needed to answer science questions.
%We also explored the impact of joint training with review questions. 
%Review questions are known to be helpful in student learning. 
We found that joint training with review questions helped improve accuracy as well.
%These results show significant value in our approach of modeling science QA via latent structures i.e. exploring portions of text in the student curriculum that align well with the question-answers, modeling negation as a special case, in handling various question types via a multi-task model, and in utilizing review questions.
%We found that our method performs better on questions without context compared to questions with context. Using question context for inference is a challenge for future work.
\\
{\bf Feature Ablation:}
As described before, our feature set comprises of five parts, where each part corresponds to a part of the answer-entailing structure -- textbook ($\z_1$), chapter ($\z_2$), section ($\z_3$), snippets ($\z_4$), and alignment ($\z_5$). 
It is interesting to know the relative importance of these parts in our model. Hence, we perform feature ablation on our best performing model - \textit{MTLSSVM(QWord, JT)} where we remove the five feature parts one by one and measure the loss in accuracy. Figure \ref{ablation-fig} shows that the choice of section and alignment are important components of our model. Yet, all components are important and removing any of them will result in a loss of accuracy. Finally, in order to understand the value of external knowledge resources (K), we removed the component that induces and aligns the hypothesis with knowledge bits. This results in significant loss in performance, estabishing the efficacy of adding in external knowledge via our approach.

%% file: conclusions.tex
\section{Conclusion}
We addressed the problem of answering $8^{th}$ grade science questions using textbooks, domain specific dictionaries and semi-structured tables. We posed the task as an extension to textual entailment and proposed a solution that learns latent structures that align question answer pairs with appropriate snippets in the textbooks. Using domain specific dictionaries and semi-structured tables, we further refined the structures. The task required handling a variety of question types so we extended our technique to multi-task setting. 
Our technique showed improvements over a number of baselines. Finally, we also used a set of associated review questions, which were used to gain further improvements.
%Our technique would benefit from domain specific dictionaries and knowledge bases, and more question specific analysis. For example, information about periodic table, chemical compounds (e.g. carbon-dioxide and CO$_2$ refer to the same compound), mathematical concepts (e.g. temperature conversion), etc. would help.